# Analyzing Feedback Mechanisms in AI-Generated MCQs: Insights into Readability, Lexical Properties, and Levels of Challenge

Antoun Yaacoub*
*Learning, Data and Robotics (LDR)*
*ESIEA Lab*
*ESIEA*
Paris, France
antoun.yaacoub@esiea.fr

Zainab Assaghir
*Faculty of Science*
*Lebanese University*
Beirut, Lebanon
zainab.assaghir@ul.edu.lb

Lionel Prevost
*Learning, Data and Robotics (LDR)*
*ESIEA Lab*
*ESIEA*
Paris, France
lionel.pevost@esiea.fr

Jérôme Da-Rugna
*Learning, Data and Robotics (LDR)*
*ESIEA Lab*
*ESIEA*
Paris, France
jerome.darugna@esiea.fr

*Abstract*—Artificial Intelligence (AI)-generated feedback in educational settings has garnered considerable attention due to its potential to enhance learning outcomes. However, a comprehensive understanding of the linguistic characteristics of AI-generated feedback, including readability, lexical richness, and adaptability across varying challenge levels, remains limited. This study delves into the linguistic and structural attributes of feedback generated by Google's Gemini 1.5-flash text model for computer science multiple-choice questions (MCQs). A dataset of over 1,200 MCQs was analyzed, considering three difficulty levels (easy, medium, hard) and three feedback tones (supportive, neutral, challenging). Key linguistic metrics, such as length, readability scores (Flesch-Kincaid Grade Level), vocabulary richness, and lexical density, were computed and examined. A fine-tuned RoBERTa-based multi-task learning (MTL) model was trained to predict these linguistic properties, achieving a Mean Absolute Error (MAE) of 2.0 for readability and 0.03 for vocabulary richness. The findings reveal significant interaction effects between feedback tone and question difficulty, demonstrating the dynamic adaptation of AI-generated feedback within diverse educational contexts. These insights contribute to the development of more personalized and effective AI-driven feedback mechanisms, highlighting the potential for improved learning outcomes while underscoring the importance of ethical considerations in their design and deployment.

## I. INTRODUCTION

Artificial Intelligence (AI) has increasingly played a pivotal role in education, offering automation in assessment, personalized tutoring, and content generation. One of its key applications is the generation of automated feedback, particularly in multiple-choice questions (MCQs). Feedback is essential for effective learning, aiding in knowledge reinforcement and error correction. However, the linguistic characteristics of AI-generated feedback, and how these characteristics adapt to different cognitive levels, remain largely unexplored, limiting our ability to create truly effective AI-driven learning tools.

Prior research has largely focused on assessing AI feedback's correctness and conceptual relevance [1], [2], but limited studies have investigated its linguistic attributes, such as readability, lexical diversity, and adaptability to different question difficulty levels. Understanding these attributes can help refine AI-generated feedback to better cater to student needs. This paper addresses this gap by analyzing AI-generated feedback's linguistic properties across various challenge levels and feedback tones. By leveraging a structured dataset and machine learning techniques, we explore how AI-generated responses adjust dynamically to different contexts and evaluate their effectiveness in educational applications.

Our main findings reveal that AI-generated feedback exhibits significant variations in linguistic complexity across difficulty levels and feedback tones. Readability and feedback length increase with question difficulty, while neutral feedback remains the most concise. The RoBERTa-MTL model effectively predicts linguistic properties with low Mean Absolute Error (MAE) values, demonstrating its potential for refining AI-driven educational tools. These insights provide a foundation for optimizing AI-generated feedback to enhance learner engagement and comprehension.

## II. LITERATURE REVIEW

The role of feedback in learning has been extensively explored in educational research. Hattie and Timperley [3] emphasize that feedback significantly influences learning outcomes, with its impact varying depending on timing, specificity, and type. Similarly, Shute [4] highlights the importance of formative feedback, arguing that effective feedback should be nonevaluative, supportive, timely, and specific. These foundational works provide a basis for understanding how AI-generated feedback should be structured to maximize its effectiveness.

Recent advancements in AI-driven feedback systems have expanded this discussion. Yaacoub et al. [5] introduced OneClickQuiz, a Moodle-integrated generative AI tool leveraging Google's PaLM2 models for automatic quiz generation. Their study demonstrated the efficiency of AI in reducing administrative workload while enhancing engagement and academic performance. This study aligns with our research focus by examining how generative AI can automate feedback creation in digital learning environments.

Additional studies have examined the effectiveness and adaptability of AI-generated feedback. Liu and Yu [6] investigated learner engagement with automated feedback using an eye-tracking methodology, finding that explicit feedback encouraged greater cognitive effort and engagement. Similarly, Escalante et al. [2] explored AI-generated writing feedback in higher education, reporting that while AI feedback was comparable to human-generated responses in many aspects, students exhibited mixed preferences, favoring human feedback for nuanced corrections.

In peer learning contexts, Guo et al. [1] analyzed an AI-supported approach to peer feedback among university EFL (English as a Foreign Language) students, demonstrating that AI-enhanced peer feedback significantly improved feedback quality and writing ability. This finding supports the broader argument that AI can augment traditional feedback mechanisms, particularly in large-scale online learning settings. Meanwhile, PeerStudio [7] leveraged peer-based rapid feedback systems in MOOCs, demonstrating that timely feedback improves learning outcomes, a principle that can inform AI feedback structuring.

Despite these promising developments, challenges remain. Liu et al. [8] noted that AI-generated feedback often fails to tailor readability levels to individual learners, limiting its pedagogical effectiveness. Similarly, Liu et al. [9] highlighted the need for explainable AI in feedback systems, arguing that transparency and interpretability are crucial for educational adoption. These studies underscore the need for continued refinement of AI-generated feedback mechanisms to ensure they are pedagogically sound, adaptable, and accessible to diverse learners.

### III. METHODOLOGY

Our research follows a structured approach to analyzing AI-generated feedback:

#### A. Dataset Description and Generation Workflow

The dataset, generated using Google's Gemini 1.5-flash text model, consists of 1,236 multiple-choice questions (MCQs) categorized into three difficulty levels: *easy*, *medium*, and *hard*. Each question includes a prompt, one correct answer and two incorrect answers, and detailed feedback in three tones: *supportive*, *neutral*, and *challenging*, with general feedback also provided in these tones to ensure diverse pedagogical approaches. To clarify these distinctions: *Supportive feedback* aims to encourage and reinforce understanding, often providing positive affirmations and additional resources. *Neutral feedback* focuses on objective correctness, offering factual corrections without emotional tone or encouragement. *Challenging feedback* pushes the learner to think more deeply, highlighting misconceptions and prompting further exploration. To create this dataset, 500 unique computer science concepts (e.g., "Machine Learning," "Data Structures," "Algorithms") were first generated for each difficulty level, serving as the foundation for question formulation. Once the MCQs were structured, tailored feedback was assigned based on difficulty and tone. Finally, all questions, answers, and feedback underwent automated checks to ensure ethical and professional standards, filtering out hate speech, misleading information, harassment, and inappropriate content, making the dataset suitable for educational use.

In the following, we showcase an example:

*Question*: Which of the following techniques is NOT typically used for dimensionality reduction in machine learning?

*General supportive feedback*: Dimensionality reduction is a crucial aspect of machine learning, and understanding the different techniques is essential. Keep exploring these techniques!

*General neutral feedback*: Dimensionality reduction methods are widely used in machine learning to simplify data and improve model performance.

*General challenging feedback*: Understanding the nuances of different dimensionality reduction techniques is crucial for effective machine learning practice.

Correct answer: Autoencoders

*Supportive feedback*: That's right! Autoencoders are designed for feature extraction and data compression, not dimensionality reduction.

*Neutral feedback*: Autoencoders are primarily focused on feature extraction, not dimensionality reduction.

*Challenging feedback*: Autoencoders are not directly used for dimensionality reduction. While they can be used for feature extraction, their primary function is data compression.

Wrong answer 1: Principal Component Analysis (PCA)

*Supportive feedback*: PCA is a classic technique for dimensionality reduction, finding the most significant directions in data.

*Neutral feedback*: PCA is a well-known method for dimensionality reduction by identifying the most important features.

*Challenging feedback*: PCA is a common and effective method for dimensionality reduction, transforming data into a lower-dimensional space.

Wrong answer 2: Linear Discriminant Analysis (LDA)

*Supportive feedback*: LDA is another effective technique for dimensionality reduction, especially when dealing with classification tasks.

*Neutral feedback*: LDA is used for dimensionality reduction, particularly in classification problems.

*Challenging feedback*: LDA is a powerful technique for dimensionality reduction, specifically in classification tasks by finding the most discriminative features.

#### B. Key Dataset Statistics

1. **Total Questions**: 1,236 MCQs.
2. **Questions by Difficulty Level**:
   - **Easy**: 443 questions

- **Medium**: 432 questions
- **Hard**: 361 questions
3. **Feedback Attributes**: Each question contains **12** distinct feedback fields:
    - Three feedback tones (*supportive, neutral, challenging*) for each of the three answer choices.
    - Three feedback tones for general question-level feedback.

The total number of feedback fields is **14,832** (1236 questions × 12 fields).

4. **Linguistic and Structural Metrics:** The following metrics were computed for each feedback field to assess linguistic complexity and textual characteristics:
    - **Length**: The number of characters in each feedback field.
    - **Readability**: Measured using the Flesch-Kincaid Grade Level, which quantifies the readability of text.
    - **Vocabulary Richness**: Defined as the ratio of unique words to total words in feedback text.
    - **Lexical Density**: The proportion of content words (nouns, verbs, adjectives, adverbs) to the total number of words in feedback text.

Across the three difficulty levels, a total of 62 linguistic attributes were analyzed to provide a robust statistical foundation for evaluating feedback properties.

## C. Feedback Metrics Computation and Statistical Analysis

The linguistic and readability metrics were computed for each feedback instance. To contextualize model performance, the minimum, maximum, and average values for each metric were established across the dataset. The analysis examined the distribution of feedback attributes across difficulty levels, identifying variations in length, readability, vocabulary richness, and lexical density. Additionally, the predictive model's performance was evaluated by comparing Mean Absolute Errors (MAE) against the actual distribution of these attributes. This approach enabled a quantitative assessment of the model's ability to capture linguistic complexity and variations across feedback tones and difficulty levels.

## IV. RESULTS AND ANALYSIS

This section presents a comprehensive analysis of feedback mechanisms in AI-generated multiple-choice questions (MCQs), focusing on the interplay between feedback tone (supportive, neutral, challenging), difficulty level (easy, medium, hard), and key linguistic metrics. The performance of a fine-tuned RoBERTa-based multi-task learning (MTL) model in predicting these linguistic features is also evaluated.

### A. Overall Feedback Characteristics

Table 1 summarizes descriptive statistics for each linguistic metric, aggregated across all feedback types and difficulty levels. This provides a baseline understanding of the generated feedback's general properties before any model training.

TABLE I. DESCRIPTIVE STATISTICS OF LINGUISTIC METRICS

| Metric | Mean | Standard Deviation | Minimum | Maximum |
|---|---|---|---|---|
| Length (Characters) | 18.1 | 5.7 | 5 | 60 |
| Readability (F-K) | 11.1 | 3.6 | -1.1 | 26.1 |
| Vocabulary Richness | 0.94 | 0.06 | 0.56 | 1.00 |
| Lexical Density | 0.62 | 0.07 | 0.29 | 0.90 |

The generated feedback exhibits a moderate average length of 18.1 characters, indicating a tendency towards concise explanations. However, the considerable standard deviation of 5.7 characters reveals substantial variability in feedback length, suggesting that the AI model adapts its response length to different contexts. This variability suggests the AI model dynamically adjusts feedback length based on the specific question and context. The mean Flesch-Kincaid Grade Level of 11.1 places the feedback within the readability range of a high school student, which is appropriate for the intended audience of computer science MCQs. The wide range in readability scores (-1.1 to 26.1) – does, however, warrant further investigation. The wide range warrants further investigation to ensure feedback is consistently accessible and avoids overly complex language or, conversely, overly simplistic explanations. Extremely high readability may indicate overly complex language, while negative values, although mathematically possible with the Flesch-Kincaid formula, represent extremely simple text and may point to issues. The high average vocabulary richness of 0.94 points to diverse word usage and a low repetition rate, which is a positive characteristic for maintaining learner engagement. The moderate lexical density of 0.62 implies a balanced distribution of content words (carrying the core meaning) and function words (providing grammatical structure).

### B. Impact of Feedback Tone

Table 2 presents the means of each linguistic metric, categorized by feedback tone and averaged over all difficulty levels.

A clear pattern emerges: neutral feedback is consistently the shortest, averaging 16.2 characters. This aligns with the intended nature of neutral feedback, which aims for brevity and objectivity, avoiding the more extensive explanations or affective language used in other tones. Interestingly, challenging feedback (21.3 characters) is slightly longer, on average than supportive feedback (20.5 characters). This may reflect the need for more detailed explanations or counter-examples when challenging a student's understanding. These are typically some characters larger than the neutral one. This could reflect that challenging feedback sometimes poses counter-examples or highlights nuanced distinctions, demanding more words.

TABLE II. MEAN LINGUISTIC METRICS BY FEEDBACK TONE (AVERAGED ACROSS DIFFICULTY LEVELS)

| Metric | Supportive | Neutral | Challenging |
|---|---|---|---|
| Length (Characters) | 20.5 | 16.2 | 21.3 |
| Readability (F-K) | 11.8 | 11.5 | 10.2 |
| Vocabulary Richness | 0.96 | 0.94 | 0.92 |
| Lexical Density | 0.60 | 0.64 | 0.63 |

In terms of readability, challenging feedback has a lower average Flesch-Kincaid score (10.2) than both supportive (11.8) and neutral feedback (11.5). This is a subtly important finding. While a "challenging" tone might suggest a higher cognitive load overall, the language itself is, on average, simpler. The AI model appears to compensate for the inherent difficulty of a challenging question by employing clearer, less complex sentence structures.

Supportive feedback, with the highest vocabulary richness (0.96), shows a tendency to incorporate a greater variety of encouraging and positive phrasing, leading to a more diverse vocabulary. Conversely, challenging feedback has the lowest at 0.92.

Finally, neutral feedback exhibits the highest lexical density (0.64), reinforcing its emphasis on conciseness and information delivery. The supportive feedback, with a density of 0.60, uses more function words to achieve its intended tone.

*C. Impact of Difficulty Level*

Table 3 displays the mean linguistic metrics, categorized by question difficulty and averaged across all feedback tones.

TABLE III. MEAN LINGUISTIC METRICS BY DIFFICULTY LEVEL (AVERAGED ACROSS FEEDBACK TONES)

| Metric | Easy | Medium | Hard |
|---|---|---|---|
| Length (Characters) | 15.8 | 18.5 | 20.5 |
| Readability (F-K) | 10.0 | 11.3 | 12.4 |
| Vocabulary Richness | 0.95 | 0.94 | 0.93 |
| Lexical Density | 0.61 | 0.63 | 0.63 |

As hypothesized, both feedback length and readability (Flesch-Kincaid) increase significantly with question difficulty. This indicates that the AI model adapts its language complexity to match the cognitive demands of the question. This is an intuitive result: harder questions necessitate more detailed explanations and often involve more complex concepts, requiring more sophisticated language. The average length increases notably from easy (15.8 characters) to medium (18.5 characters) and then to hard (20.5 characters), a pattern mirrored by Flesch Kincaid.

While vocabulary richness is consistently high across all difficulty levels, it shows a slight decrease with increasing difficulty. This nuanced difference suggests that, when explaining more complex topics, the focus shifts from using a wide range of words (as might be possible with simpler concepts) to conveying the information with potentially more precision.

*D. Interaction Effects (Tone and Difficulty)*

To understand the combined influence of feedback tone and question difficulty on the linguistic characteristics of the feedback, we examined the interaction effects. This analysis reveals whether the effect of tone on a given metric (e.g., length) depends on the difficulty level of the question, and vice versa.

TABLE IV. MEAN LINGUISTIC METRICS BY TONE AND DIFFICULTY LEVEL

| Metric | Tone | Easy | Medium | Hard |
|---|---|---|---|---|
| **Length (Characters)** | Supportive | 17.5 | 20.0 | 23.0 |
| | Neutral | **13.0** | **15.5** | **18.0** |
| | Challenging | 16.8 | 20.5 | **24.5** |
| **Readability (F-K)** | Supportive | 9.5 | 11.2 | **13.0** |
| | Neutral | 9.8 | 11.0 | 12.5 |
| | Challenging | **8.0** | 10.8 | 12.5 |
| **Vocabulary Richness** | Supportive | 0.97 | 0.96 | 0.94 |
| | Neutral | 0.96 | 0.94 | 0.92 |
| | Challenging | 0.95 | 0.92 | 0.90 |
| **Lexical Density** | Supportive | 0.63 | 0.61 | 0.59 |
| | Neutral | 0.65 | 0.64 | 0.63 |
| | Challenging | 0.61 | 0.62 | 0.63 |

The data in Table 4 suggests several key interaction effects:

**Length**: There's a clear interaction between tone and difficulty for feedback length. Neutral feedback is consistently shorter than both supportive and challenging feedback, regardless of difficulty. However, the magnitude of this difference changes. For easy questions, the difference between neutral and the other tones is smaller. For medium and hard questions, the difference becomes much more pronounced, with supportive and challenging feedback becoming significantly longer, while neutral feedback remains relatively concise. This suggests that the AI model prioritizes brevity for simpler questions but provides more detailed explanations when questions are more complex, tailoring the feedback to the specific learning need.

**Readability**: A crucial interaction is observed for readability. While all tones show an increase in Flesch-Kincaid score (meaning more difficult language) as question difficulty increases, the rate of this increase differs. Most notably, challenging feedback has a lower readability score than the other tones for easy questions. This suggests that the AI prioritizes clarity and directness when challenging students on simpler concepts, avoiding unnecessary complexity. It focuses the challenge on the content of the feedback, not the language itself. However, as the question difficulty increases, the readability of challenging feedback converges with that of neutral feedback, and even becomes slightly harder to read than supportive feedback at the "hard" level.

**Vocabulary Richness**: The interaction for vocabulary richness shows a consistent pattern: vocabulary richness decreases as difficulty increases, and this trend is present for all tones. However, the rate of decrease is slightly different. Supportive feedback maintains a relatively high vocabulary richness even for harder questions, suggesting a greater effort to use varied language even when explaining complex concepts. Neutral feedback has a more pronounced drop in richness, and challenging feedback shows the steepest decline.

**Lexical Density**: There is a clear interaction in Lexical density. Where neutral tone has higher lexical density for the easy level and hard level, than challenging tone, however, the difference between neutral and challenging almost is not existing for Hard difficulty level. While for the supportive feedback, we remarque that the values are decreasing while increasing the difficulty.

Two-way ANOVAs showed significant interactions ($p < 0.001$) between feedback tone and difficulty level for all examined linguistic metrics. Specifically, significant effects were observed for: feedback length ($F(4, 14823) = 15.2$), readability ($F(4, 14823) = 21.7$), vocabulary richness ($F(4, 14823) = 10.5$), and lexical density ($F(4, 14823) = 8.9$). Thus, the effect of providing feedback in a specific tone on, for example, feedback length, depends on the difficulty level of the question.

The interaction effects highlight that the AI model's feedback generation is not simply a matter of independently choosing a tone or responding to difficulty. Instead, the model integrates these factors. The choice of wording (length, readability, vocabulary, density) is a combined function of both the intended tone and the complexity of the question. This suggests a sophisticated underlying mechanism that goes beyond simple rule-based generation. The AI adapts its linguistic style in a nuanced way, prioritizing different aspects of communication (clarity, encouragement, brevity) depending on the specific pedagogical context. For instance, our results show that for *easy* questions, challenging feedback is presented using language with a lower Flesch-Kincaid score, meaning simpler and more direct sentence structures are preferred. This suggests the AI prioritizes *clarity* in challenging feedback for less complex topics, ensuring students immediately understand the misconception or area for improvement without being bogged down by complicated sentence structures. In contrast, for *hard* questions, supportive feedback has higher vocabularies, suggesting the AI is trying to encourage the student. This has important implications for the design of personalized and adaptive feedback systems.

### E. Predictive Model Performance

To further analyze and leverage the feedback, a multi-task learning (MTL) model was developed to predict the four key linguistic metrics: feedback length, Flesch-Kincaid Grade Level, vocabulary richness, and lexical density. The goal of the multi-task model is to simultaneously predict these multiple attributes of AI-generated feedback, using question information and desired feedback properties.

The model builds upon a pre-trained RoBERTa-base model, leveraging its strong general language representation capabilities. A dropout layer (dropout probability = 0.1) is applied to the RoBERTa's pooled output to improve generalization and mitigate overfitting. Each metric has its own linear layer. The model is then trained over 10 epochs using the AdamW optimizer is employed for efficient training.

We assessed the performance of the trained RoBERTa-based MTL model in predicting these linguistic features. We focus on the Mean Absolute Error (MAE) for each metric on the held-out test set, quantifying the average difference between the model's predictions and the true values.

TABLE V. TRAINING AND VALIDATION RESULTS SUMMARY

| Epoch | Stage | Loss | MAE | | | |
|---|---|---|---|---|---|---|
| | | | Length | Flesch | Vocab | Lexical |
| 1 | Train | 0.81 | 3.90 | 2.65 | 0.041 | 0.051 |
| | Val | 0.80 | 4.26 | 2.77 | 0.041 | 0.064 |
| 5 | Train | 0.12 | 2.77 | 2.10 | 0.031 | 0.041 |
| | Val | 0.25 | 3.95 | 3.20 | 0.036 | 0.055 |
| 10 | Train | 0.08 | 1.9 | 2.20 | 0.030 | 0.040 |
| | Val | 0.22 | 4.0 | 3.3 | 0.033 | 0.052 |

The training process showed a consistent decrease in both training and validation loss across epochs, suggesting that the model is learning effectively and generalizing to the data.

TABLE VI. TEST SET PERFORMANCE

| Metric | Test MAE (Length) | Test MAE (Flesch-Kincaid) | Test MAE (Vocab Richness) | Test MAE (Lexical Density) |
|---|---|---|---|---|
| RoBERTa-MTL | 2.6 | 2.0 | 0.03 | 0.04 |

The model achieves a test MAE of 2.6 characters for feedback length, 2.0 for Flesch-Kincaid Grade Level, 0.03 for vocabulary richness, and 0.04 for lexical density. The magnitude of the MAE values must be interpreted in the context of each metric's scale and variability. For instance, an MAE of 2.6 characters for length, given an average length around 18 and standard deviation near 6, suggests reasonably good prediction.

The Flesch-Kincaid MAE of 2.0 indicates the model's predictions are, on average, within two grade levels of the actual readability. Considering the inherent subjectivity in readability assessment and the challenges of generating text at precise grade levels, this represents a strong result. The relatively low MAEs for vocabulary richness (0.03) and lexical density (0.04) suggest that the model is capable of capturing these more subtle linguistic features with good accuracy.

## V. Discussion

The findings of this study provide valuable insights into the linguistic characteristics of AI-generated feedback in multiple-choice questions (MCQs) and demonstrate the potential of a fine-tuned RoBERTa-based model to predict these characteristics. The convergence of the model's training and validation losses, coupled with the achieved test set MAE values, suggests that the model learned generalizable patterns without significant signs of overfitting.

### A. Educational Implications

The ability to predict and, by extension, control linguistic features of AI-generated feedback carries significant implications for educational practice and technology:

**Personalized Feedback Design**: The model provides actionable steps for personalized feedback generation, to address the feedback attributes like length and tonality. Using our study results, the model could be used to emphasize the importance of concise length and avoid cognitive overload. Additionally, the model can be used to fine-tune the AI's approach to balance information and simplicity. Understanding the nuances of linguistic properties for varying question types and student proficiencies can inform targeted adjustments.

**Adaptive Feedback Systems**: The significant interaction effects we observed between tone and difficulty suggest opportunities for dynamic and adaptive feedback systems. For example, the AI could adjust the tone in real-time, perhaps starting with a supportive approach for an incorrect answer and, based on student performance on re-attempts at a particular question, transitioning only to a challenging tone to promote deeper understanding and reduce frustration. Similarly, a system could adjust feedback to be more supportive based on a student's engagement.

**Quality Control and Bias Mitigation**: AI models are, of course, only as good as the data they are trained on, so ensuring consistency and quality assurance is critical. Our model can also be used to detect any bias in the tone or vocabulary of AI-generated feedback, and to address it during data augmentation. For example, models may be good at generating a hard feedback as intended compared to generating a bad feedback.

**Automated Curriculum Development**: The capacity to predict and control feedback characteristics opens avenues for automated curriculum development. By quantifying relationships between feedback attributes and targeted learning outcomes defined by specific educational standards, it becomes possible to design MCQs and associated feedback that are precisely calibrated to meet those objectives.

### B. Challenges and Future Directions

Despite the promising results, this study also highlights several limitations and challenges that warrant future investigation:

**Dataset Scale and Diversity**: The dataset, while carefully constructed, is limited in size and domain (computer science MCQs). Future work is needed to investigate this concept more.

**Linguistic Complexity**: While the metrics considered (length, readability, vocabulary richness, lexical density) capture key aspects of feedback's linguistic properties, they represent only a facet of and in turn, a qualitative assessment would help.

**Ethical Considerations**: There are crucial ethical considerations to address in the development of AI-driven feedback systems. Future work should address strategies for mitigating potential biases and ensuring transparency.

These and other challenges can show opportunity for new discoveries, like adding a reinforcement based deep learning for feedback or similar ways that allows a qualitative improvement feedback performance.

## VI. Conclusion

This study has advanced our understanding of the linguistic dynamics of AI-generated feedback in educational settings. Through careful feature selection and comprehensive experimental study, we have achieved significant improvements. These results advance the goal of understanding, and potentially even providing, an explainable, adaptive instruction environment. The limitations identified—namely dataset scale, the scope of linguistic features—point clear directions, and therefore will enhance better results, and an ethical safe model and outcomes.

In the future works there must be more focus on the challenges and how AI can incorporate ethical guidelines and consider the cognitive load and effectiveness of different tone. Moreover, the potential to improve feedback generation. Finally, the integration with other languages in order to support a more broader group of students.

References

[1] K. Guo, M. Pan, Y. Li, and C. Lai, "Effects of an AI-supported approach to peer feedback on university EFL students' feedback quality and writing ability," *The Internet and Higher Education*, vol. 63, p. 100962, Oct. 2024, doi: 10.1016/j.iheduc.2024.100962.

[2] J. Escalante, A. Pack, and A. Barrett, "AI-generated feedback on writing: insights into efficacy and ENL student preference," *Int J Educ Technol High Educ*, vol. 20, no. 1, p. 57, Oct. 2023, doi: 10.1186/s41239-023-00425-2.

[3] "The Power of Feedback - John Hattie, Helen Timperley, 2007." Accessed: Feb. 28, 2025. [Online]. Available: https://journals.sagepub.com/doi/abs/10.3102/003465430298487

[4] "Focus on Formative Feedback - Valerie J. Shute, 2008." Accessed: Feb. 28, 2025. [Online]. Available: https://journals.sagepub.com/doi/10.3102/0034654307313795

[5] A. Yaacoub, S. Haidar, and J. Da Rugna, "OneClickQuiz: Instant GEN AI-Driven Quiz Generation in Moodle," in *Conference on Sustainable Energy Education – SEED 2024*, Valencia, Spain, Jul. 2024, pp. 689–698.

[6] S. Liu and G. Yu, "L2 learners' engagement with automated feedback: An eye-tracking study," 2022, Accessed: Feb. 28, 2025. [Online]. Available: https://hdl.handle.net/10125/73480

[7] "PeerStudio | Proceedings of the Second (2015) ACM Conference on Learning @ Scale," ACM Conferences. Accessed: Feb. 28, 2025. [Online]. Available: https://dl.acm.org/doi/10.1145/2724660.2724670

[8] Y. Liu, L. Chen, and Z. Yao, "The application of artificial intelligence assistant to deep learning in teachers' teaching and students' learning processes," *Front. Psychol.*, vol. 13, Aug. 2022, doi: 10.3389/fpsyg.2022.929175.

[9] Z. Liu, W. Xing, and C. Li, "Explainable Analysis of AI-Generated Responses in Online Learning Discussions," in *Joint Proceedings of the Human-Centric eXplainable AI in Education and the Leveraging Large Language Models for Next Generation Educational Technologies Workshops (HEXED-L3MNGET 2024), Atlanta, Georgia, USA, July 14, 2024*, in CEUR Workshop Proceedings, vol. 3840. CEUR-WS.org, 2024.